\documentclass[journal]{IEEEtran}

\usepackage{graphicx}
\usepackage{amssymb}
\usepackage{booktabs}
\usepackage{multirow}
\usepackage{caption}
\usepackage{color,xcolor}
\usepackage{algorithm}
\usepackage{algorithmic}
\usepackage{animate}
\usepackage{marvosym}
\usepackage{bm}
\usepackage{amsmath}
\usepackage[pagebackref=true,breaklinks=true,colorlinks,bookmarks=false]{hyperref}

\ifCLASSOPTIONcompsoc
\usepackage[nocompress]{cite}
\else
\usepackage{cite}
\fi

\begin{document}
\title{{\LARGE Cross-modal Local Shortest Path and Global Enhancement\\
			 for Visible-Thermal Person Re-Identification}}

%%*************************************************************************

\author{Xiaohong Wang\textsuperscript{\Letter},
	 Chaoqi Li,
	and~Xiangcai Ma
	\thanks{Xiaohong Wang with the University of Shanghai for Science and Technology , Shanghai,30332 China E-mail:wangxiaohong@st.usst.edu.cn}% <-this % stops a space
	\thanks{ Chaoqi Li and Xiangcai Ma are with the University of Shanghai for Science and Technology.}% <-this % stops a space
	}
 
% The paper headers
\markboth{Journal of \LaTeX\ Class Files,~Vol.~14, No.~8, August~2015}%
{Shell \MakeLowercase{\textit{et al.}}: Bare Demo of IEEEtran.cls for Computer Society Journals}

%%*************************************************************************
\IEEEtitleabstractindextext{%
	\begin{abstract}
		In addition to considering the recognition difficulty caused by human posture and occlusion, it is also necessary to solve the modal differences caused by different imaging systems in the Visible-Thermal cross-modal person re-identification (VT-ReID) task. In this paper,we propose the Cross-modal Local Shortest Path and Global Enhancement    (CM-LSP-GE) modules, a two-stream network based on joint learning of local and global features. The core idea of our paper is to use local feature alignment to solve occlusion problem, and to solve modal difference by strengthening global feature. Firstly, Attention-based two-stream ResNet network is designed to extract dual-modality features and map to a unified feature space. Then, to solve the cross-modal person pose and occlusion problems, the image are cut horizontally into several equal parts to obtain local features and the shortest path in local features between two graphs is used to achieve the fine-grained local feature alignment. Thirdly, a batch normalization enhancement module applies global features to enhance strategy, resulting in difference enhancement between different classes. The multi granularity loss fusion strategy further improves the performance of the algorithm. Finally, joint learning mechanism of local and global features is used to improve cross-modal person re-identification accuracy. The experimental results on two typical datasets show that our model is obviously superior to the most state-of-the-art methods. Especially, on SYSU-MM01 datasets, our model can achieve a gain of 2.89$\%$and 7.96$\%$in all search term of Rank-1 and mAP. The source code will be released soon.
	\end{abstract}
	
	% Note that keywords are not normally used for peerreview papers.
	}

% make the title area
\maketitle
\IEEEdisplaynontitleabstractindextext
\IEEEpeerreviewmaketitle

%---------------------------------------------------- ---------------------
\section{Introduction}
\label{sec:introduction}
	\IEEEPARstart{P}{erson} re-identification(Re-ID)\textsuperscript{\cite{ye2021deep}}is an important technology in video tracking, which mainly studies matching person images from different camera perspectives. The traditional Re-ID mainly focuses on the visible modality, but in the actual application environment, there are often problems that cannot be identified due to low illumination, so that full-time tracking cannot be realized. In order to obtain the information of pedestrian identification at night, researchers turned to VT-Reid\textsuperscript{\cite{wu2017rgb}}research which mainly studies the matching of pedestrian images from one modality (infrared) to another modality (visible), which is a cross-modal tracking technology for intelligent tracking of pedestrians in full time. 
	
\begin{figure}[!t]
		\centering
		\includegraphics[width=0.95\linewidth]{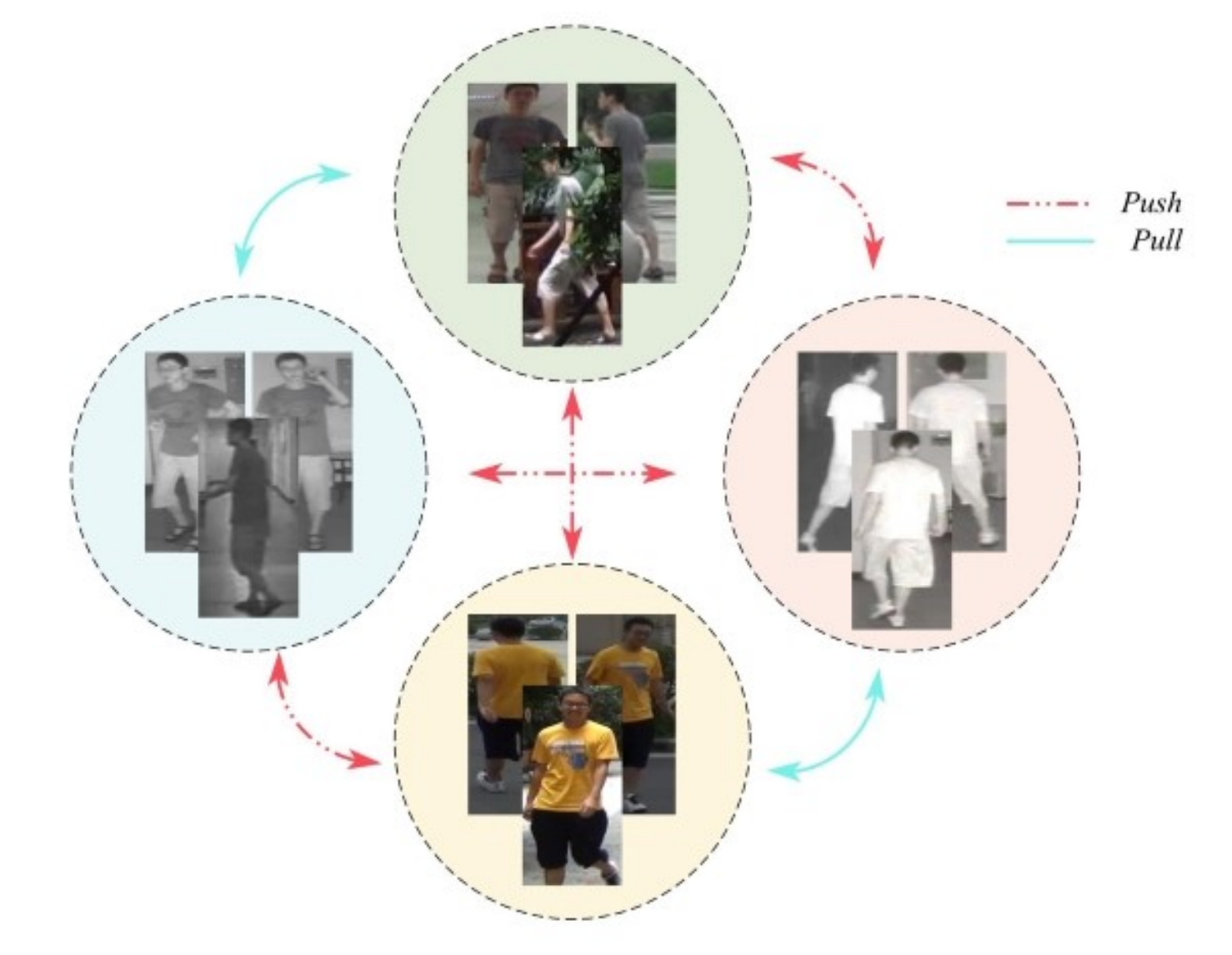}
		\vspace{-.05in}
		\caption{The cross-modal person re-identification technology difficulties. Sample images from SYSU-MM01 dataset\textsuperscript{\cite{wu2017rgb}}}
		\label{fig challenge}
	\end{figure}

	As shown in Figure~\ref{fig challenge}, now the main challenge of VT-Reid is not only to solve the problem of pedestrian posture difference and body part occlusion, but also to solve the problem of image inconsistency caused by modal difference.	
	%%% Figure 1 difficulties
	
	In order to improve the accuracy of cross modal pedestrian re-recognition, researchers have proposed machine learning models that based on hand-designed feature and deep learning network. Because manual features can only represent limited pedestrian low-level features which are only part of all features , the machine learning methods such as HOG\textsuperscript{\cite{dalal2005histograms}},LOMO\textsuperscript{\cite{liao2015person}} cannot fulfill the re-identification task satisfactorily. At present, deep learning is widely used to solve the re-identification problem and its common solution is to encode the pedestrian characteristics of different modes into a common feature space for similarity measurement.
	
	To alleviate the modal differences in cross matching, in this paper we use the attention-based two-stream ResNet network to extract the features of pedestrians in visible and infrared images, and map them to the unified feature space for similarity measurement. 
	
	Most of the work only focuses on the global coarse grained feature extraction process and ignore the research on the feature enhancement method after extraction, in this paper we design a feature enhancement module based on batch normalization to enhance the global feature after extraction and improve the feature representation of different pedestrians.
	
	However, it is not enough to focus only on the global features, local features are also play an important role in VT-ReID task. When the body parts are missed due to pedestrian occlusion, it is difficult to extract the global features from these images and truly characterize this person, which is easy to lead incorrect classification. Considering that local information(\textit{e.g.,} head, body)  of pedestrians in the images can be well distinguished and aids global feature learning, so in this paper, the pedestrian images under two different modes are segmented equally in the horizontal direction, and then the shortest path algorithm is used to achieve cross-modal local feature alignment. Finally, the joint learning mechanism based on local and global features can effectively improve the algorithm performance.
	
	Finally, different backbone networks in the classification task can affect the final classification accuracy. In this context, we investigate the impact of different variants of the ResNet two-stream feature extraction network on the final identification accuracy to further promote the performance of the network. 
	In summary, the contributions of this paper are:
	
	\begin{itemize}
		 % \vspace{-.1in}
		\item We propose an attention-based two-stream ResNet network for VT cross-modal feature acquisition.
		\item We propose a method for cross-modal local feature alignment based on the shortest path (CM-LSP), which effectively solves the occlusion problem in cross-modal pedestrian re-identification and improves the robustness of the algorithm.
		\item We design a batch normalized global feature enhancement (BN-GE) method to solve the problem of insufficient global feature discrimination and propose a multi granularity loss fusion strategy to guide network learning.
		\item Ours method achieves preferably results on datasets SYSU-MM01 and RegDB. This can be used as a research baseline to improve the quality of future research.
		
	\end{itemize}
%-------------------------------------------------------------------------
\section{Related Work}
	In order to improve cross-modal recognition accuracy in VT-ReID not only needs to solve the problem of pedestrian posture and occlusion, but also needs to break through the dilemma of cross-modal discrepancy. Machine learning based on artificial features has proved poor performance since they represent only some low-level pedestrian features. Therefore, Researchers turn to more powerful methods, Deep Learning, for feature acquisition and the mainstream approaches include: image generation-based methods, feature extractor-based methods and metric learning-based methods.
	
	\textbf{Image Generation Based Methods} fulfil the re-identification task by generating fake images through generative adversarial network (GAN) to reduce the difference between cross-patterns from the image level. Vladimir\textsuperscript{\cite{kniaz2018thermalgan}}et al.firstly proposed ThermalGAN to transform visible images into infrared images and then accomplish pedestrian recognition in the infrared modality. Zhang\textsuperscript{\cite{zhang2021rgb}}et al.considered a teacher-student GAN model (TS-GAN) which used modal transitions to better guide the learning of discriminative features. Xia\textsuperscript{\cite{xia2021visible}}et al.pointed out an image modal panning network which performed image modal transformation through a cycle consistency adversarial network. The above methods all use GAN to generate fake images reducing cross-modal differences from the image level. However, multiple seemingly reasonable images may be generated due to the change of the color attribute of the pedestrian appearance. It is difficult to determine which generated target is correct,resulting in a false identification process. The methods based on image generation often have the problem of algorithm performance uncertainty.  
	
	\textbf{Feature Extractor Based Methods} are mainly used to extract the distinction and consistency characteristics from different modes according to the discrepancy of different modes. Therefore, the extraction of rich features is the key of the algorithm.Wu\textsuperscript{\cite{wu2017rgb}}et al.analyzed the performance of different network structures, which include one-stream and two-stream networks, and proposed deep zero-padding for training one-stream network towards automatically evolving domain-specific nodes in the network for cross-modality matching. Kang\textsuperscript{\cite{kang2019person}} et al.rendered a one-stream model that placed visible and infrared images in different channels or created input images by connecting different channels. Fan\textsuperscript{\cite{fan2020cross}}et al.advanced a cross-spectral bi-subspace matching one-stream model to solve the matching difference between cross-modal classes. All three of the above single-stream algorithms have low accuracy due to the single-stream network structure defects, which can only extract some common features and cannot extract the discriminative features in dual-modality. 
	
	On the contrary to single stream network, the two-stream can extract different modal features by using the parallel network, so that the network has the advantage of extracting distinguishing features. Ye\textsuperscript{\cite{ye2018hierarchical}} et al. applied the two-stream AlextNet network to gain the dual-mode features, and then project these features into the public feature space. Based on this, Jiang\textsuperscript{\cite{jiang2020cross}}et al. designed a multi-granularity attention network to extract coarse-grained features separately. Ran\textsuperscript{\cite{ran2021improving}}et al. mapped global features to the same feature space and added local discriminative feature learningand the algorithms performance were improved to some extent.To verify the effect of the network flow structure to acquire features ability on the performance of the algorithm, Emrah Basaran\textsuperscript{\cite{basaran2020efficient}}designed a four-stream ResNet network framework which composed of gray flow and LZM upon the two-stream network. However, the experiment result showed that this method has large amount of calculation, high training cost and unsatisfactory experimental results. Overall, the two-stream network performs best in the structure of VT-ReID tasks.
	
	Most of the above methods use ResNet as the feature extraction network. However, the variants of ResNet, like SE\textsuperscript{\cite{hu2018squeeze}},  CBAM\textsuperscript{\cite{woo2018cbam}}, GC\textsuperscript{\cite{cao2019gcnet}}, SK\textsuperscript{\cite{li2019selective}}, ST\textsuperscript{\cite{zhang2020resnest}}, NAM\textsuperscript{\cite{liu2021nam}}, ResNetXT\textsuperscript{\cite{xie2017aggregated}}and SN\textsuperscript{\cite{luo2018differentiable}}are widely used in classification and recognition tasks, and have achieved good accuracy improvement.
	
	%%% Figure framework
	\begin{figure*}[ht]
		\centering
		\includegraphics[width=0.95\linewidth]{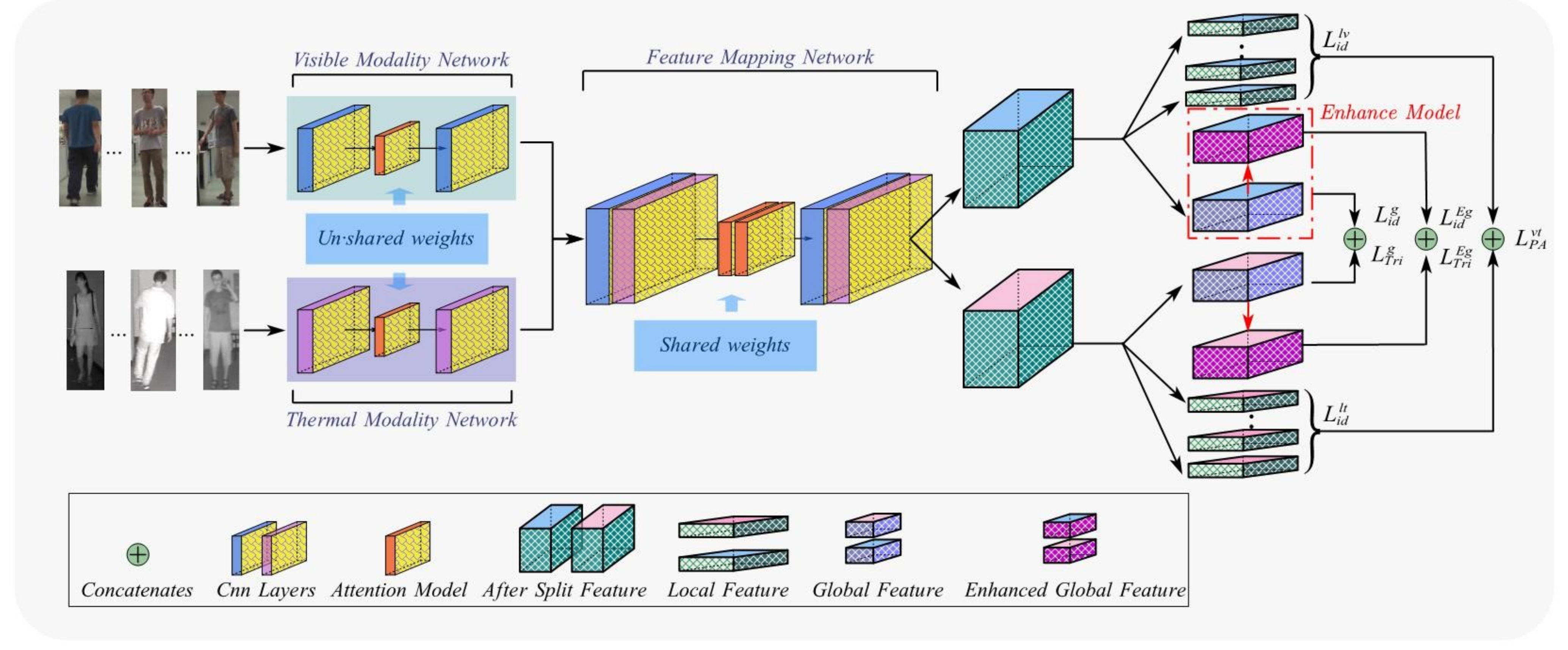}
		\caption{The model proposed in this paper consists of three main components, an attention-based two-stream backbone network, the cross-modal local feature alignment module and a global feature enhancement module. Firstly, attention-based two-stream networks extract bimodal features and map them to the same feature space. Segmentation by FMN output features to obtain unimodal After Split Feature, the segmented features continue to be divided into unimodal global features and local features containing a certain number of horizontal cuts.}
		\label{fig framework}
	\end{figure*}
	
	\textbf{Metric Learning Based Methods} are mainly focus on forcibly shortening the distance between similar samples across models and widening the distance between different samples by designing loss function. Based on the hierarchical feature extraction of the two-stream network, Ye\textsuperscript{\cite{ye2020bi,ye2018visible}}et al. designed the dual-constrained top-ranking loss and the Bi-directional exponential angular triplet loss from the global feature. Zhu\textsuperscript{\cite{zhu2020hetero}}et al.proposed to use the Hetero-center loss to constrain the intra-class center distance between two heterogeneous modes to monitor the learning of cross-modal invariant information from the perspective of global features.Ling\textsuperscript{\cite{ling2020class}}et al.advanced a center-guided metric learning method for enforcing the distance constraints among cross-modal class centers and samples. Liu\textsuperscript{\cite{liu2020enhancing}}et al.raised a dual-modality triplet loss which considering both inter-mode difference and intra-mode change and introduced a mid-level feature fusion module. Hao\textsuperscript{\cite{hao2019hsme}}et al.projected an end-to-end two-stream hypersphere manifold embedding network with classification and identification losses, which constrained the intra-mode change and cross-mode change on the hypersphere. Zhao\textsuperscript{\cite{zhao2019hpiln}}et al. introduced difficult sample quintuple loss which is used to guide global feature learning. Liu\textsuperscript{\cite{liu2021strong,liu2020parameter}}et al. introduced heterogeneous center-based triple loss and dual-granularity triple loss from cross-modal global feature alignment, and coarse-grained feature learning as well as part-level feature extraction block. Ling\textsuperscript{\cite{ling2022cross}}et al. designed the Earth Mover’s Distance can alleviate the impact of the intra-identity variations during modality alignment, and the Multi-Granularity Structure is designed to enable it to align modalities from both coarse-and fine-grained features. In order to find the nuances features, Wu\textsuperscript{\cite{wu2021discover}}et al. proposed the center clustering loss, separation loss and the mode alignment module to find the nuances of different modes in an unsupervised manner. 
	
	From the above literature, it is known that coarse-grained global feature plays a major role in recognition, while fine-grained local feature is a very good help in addition to global features to improve the ReID accuracy. In order to solve the problem of occlusion and modal difference in cross-modal VT-ReID, we use multi-granularity fusion loss to guide network learning. Firstly, in order to solve the modal difference, we design the classification loss based on global features and the loss of hard sample triples from the overall level. At the same time, the classification loss based on local features is designed from the partial level. Secondly, in order to solve the problem of component occlusion and ensure the shortest distance between components of the same kind, a part alignment loss is devised. Finally, multi-granularity loss is used to jointly constrain feature learning.	

\section{Methodology}
\label{section3}
\textbf{Overview}. In this section, we propose a CM-LSP-GE method. As shown in Fig.~\ref{fig framework}.this model mainly consists of four parts:(1) Attention-based two-stream network includes thermal mode network (TMN), visible mode network (VMN)and fusion module network (FMN). (2) Cross-modal local feature alignment module.(3) Batch normalized global feature enhancement module. (4) The multi-granularity fusion loss.	

\subsection{Attention-based two-stream network}
\label{sec:basic}
In cross-modal person re-identification project, the two-stream network is often used for feature extraction due to its excellent characteristics of discriminative learning of different modal features, 
which mainly include feature extraction and feature mapping. At present, most of the mainstream methods use ResNet50 as the backbone for feature extraction. However, some of the latest improved ResNet have achieved better performance in image classification, recognition and so on. Therefore, it is necessary to find a better backbone for VT-Reid task, and provide a new reference for future research.
\vspace{-0.8cm}
\begin{figure}[ht]
	\centering
	\includegraphics[width=0.95\linewidth]{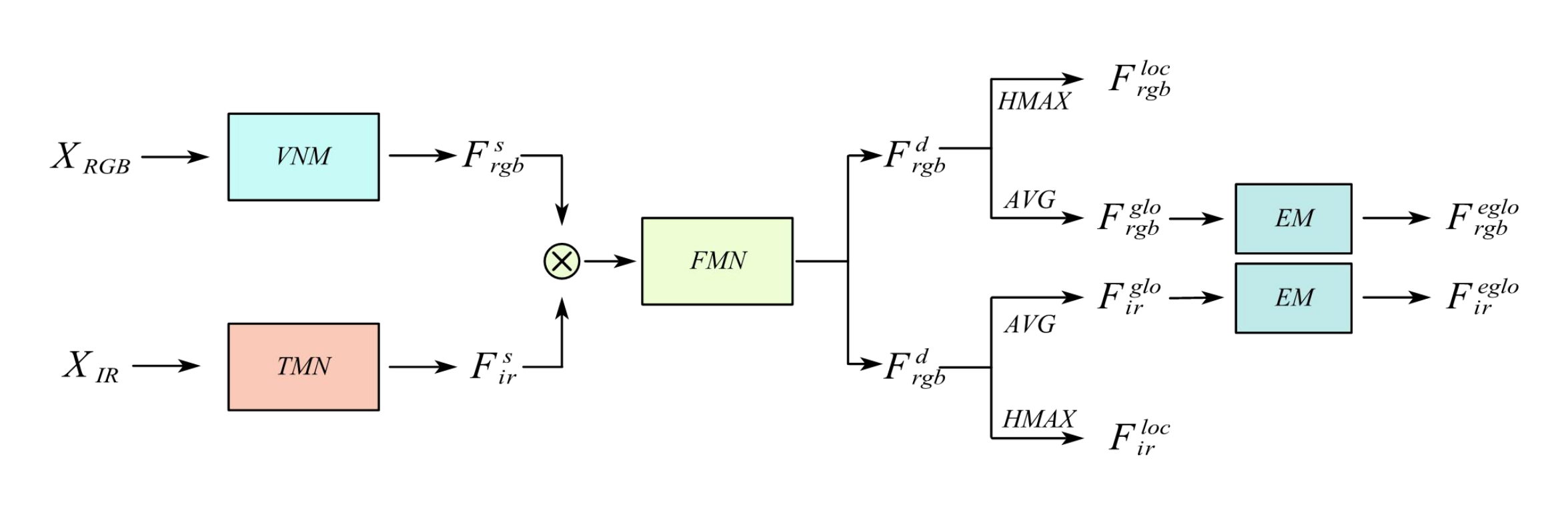}	
	
	\caption{Schematic diagram of two-stream characteristic flow}
	\label{fig3}
\vspace{-.05in}
\end{figure}

As shown in Fig.~\ref{fig3}, We will introduce how to obtain the feature of sample. Assume that the visible and infrared image are respectively defined as ${X_{rgb}}$ and ${X_{ir}}$, they are respectively fed into the VMN and TMN networks for feature learning to obtain shallow features $F_{rgb}^s$ and $F_{ir}^s$,then the two shallow features are connected as new features, which are input into the FMN network for fusion feature learning and feature mapping , and then the segmented features $F_{rgb}^d$ and $F_{ir}^d$ represented in a uniform feature space are obtained. The segmentation features are processed by the adaptive average pooling layer to obtain the global features$F_{rgb}^{glo}$ , $F_{ir}^{glo}$,while the local features$F_{rgb}^{loc}$ and $F_{ir}^{loc}$are obtained via the horizontal adaptive maximum pooling layer. Finally, the global features are enhanced by the EM module to get the enhanced features$F_{rgb}^{eglo}$ and $F_{ir}^{eglo}$.In the network inference stage, the distance matrix is constructed using the $F_{rgb}^{eglo}$ for similarity analysis.

\subsection{Cross-modal Local Feature Alignment Module}
When pedestrian occlusion occurs, it is difficult to truly recognize the global pedestrian features using missing components as distinguishing features. Therefore, we designed the shortest path alignment module based on local features, which achieves component alignment by equalizing the image segmentation and calculating the shortest path between local features in the two graphs.
\vspace{-0.8cm}
\begin{figure}[ht]
\centering
\includegraphics[width=0.95\linewidth]{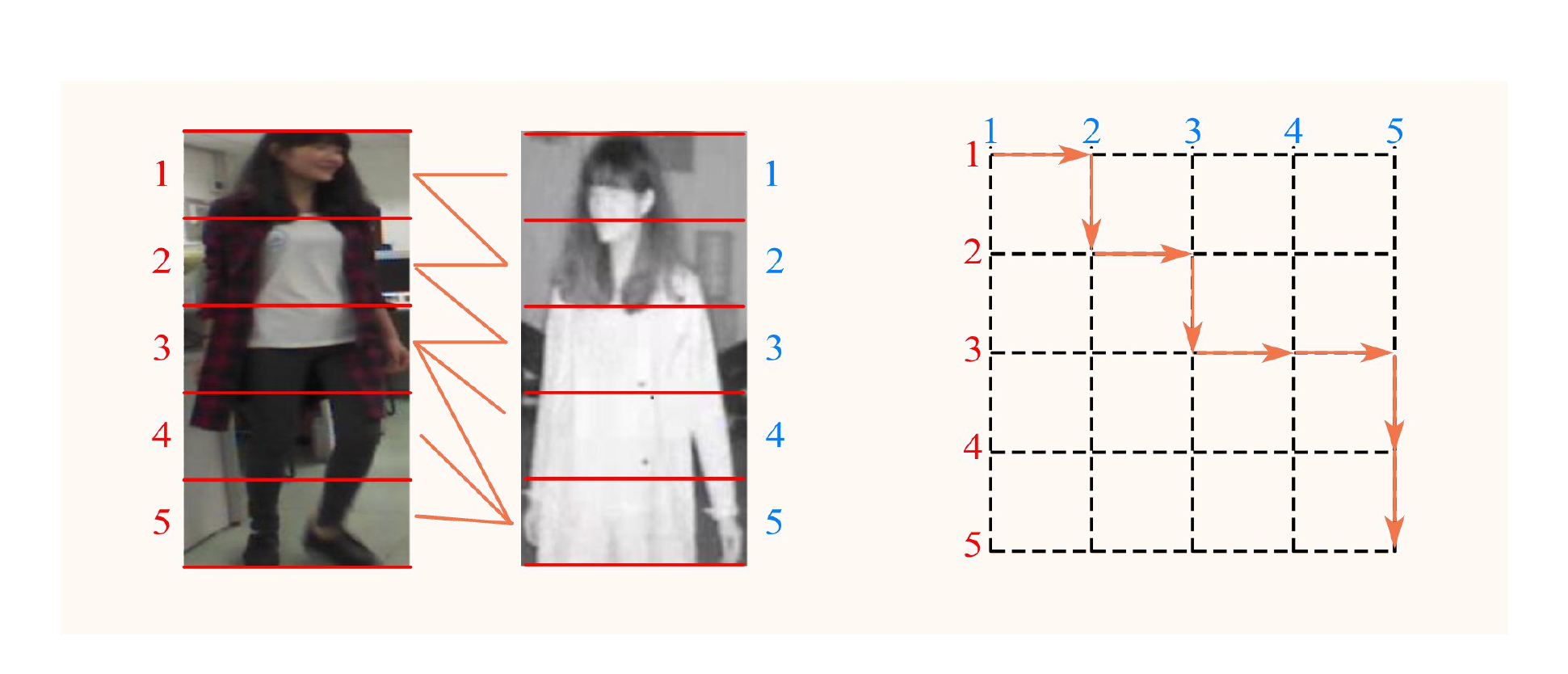}
\vspace{-.05in}
\caption{Cross-modal Local Feature Alignment}
\label{fig4}
\end{figure}

As shown in Fig.~\ref{fig4}, the visible and infrared images are equally divided into five parts, then the local feature representation is defined as $F_{rgb}^{{\rm{loc}}} = \{ f_r^1,f_r^2...f_r^i\}$ and $F_{ir}^{{\rm{loc}}} = \{ f_t^1,f_t^2...f_t^i\}$,where the visible local features are calculated as Eq.~\ref{eq:1}
	\begin{equation}
	\small
	\label{eq:1}
	f_r^i = \mathop {HMax}\limits_{i \in (1,2,...,h)} {(F_{rgb}^d)_{{\rm{[}}i \times {\rm{d]}}}}{\rm{    }}d = {2^n}\& n \in (0,1,2,...,11)
	\end{equation}
	where $i$ is the horizontal feature position, $d$ is the local feature dimension, $h$ is height of the input image.
	
	When using similar methods to calculate infrared local features, we obtain the local feature representation of the bimodal state and subsequently define the Eq.~\ref{eq:2}for calculating the distance between the two graphs. The distance equation is defined as:
	\begin{equation}
		\small
		\label{eq:2}
		{d_{i,j}} = {\left\| {\frac{{f_r^i - Mean(f_r^i)}}{{Max(f_r^i) - Min(f_r^i)}} - \frac{{f_{\rm{t}}^j - Mean(f_t^j)}}{{Max(f_t^j) - Min(f_t^j)}}} \right\|_1}
	\end{equation}
	where $i,j\in (1,2,3,...,h)$ are the respective parts of the images. ${d_{i,j}}$ is the distance between local features of different modes. 
	
	Then, we construct the distance matrix $D$ from ${d_{i,j}}$ ,and define ${S_{i,j}}$  is the total distance between the local features of the two images as the shortest distance from $(1,1)$ to $(H,H)$.the shortest path between two graphs is calculated by the Eq.~\ref{eq:3}
	\begin{equation}
	\small
	\label{eq:3}
	{S_{i,j}} = \left\{ {\begin{array}{*{20}{l}}
			{{d_{i,j}}}&{i = 1,j = 1}\\
			{{S_{i,j - 1}} + {d_{i,j}}}&{i = 1,j \ne 1}\\
			{{S_{i - 1,j}} + {d_{i,j}}}&{i \ne 1,j = 1}\\
			{\min ({S_{i,j - 1}},{S_{i - 1,j}}) + {d_{i,j}}}&{i \ne 1,j \ne 1}
	\end{array}} \right.
	\end{equation}

\subsection{Batch normalized global feature enhancement module}
\begin{figure}[!t]
		\centering
		\includegraphics[width=0.50\linewidth]{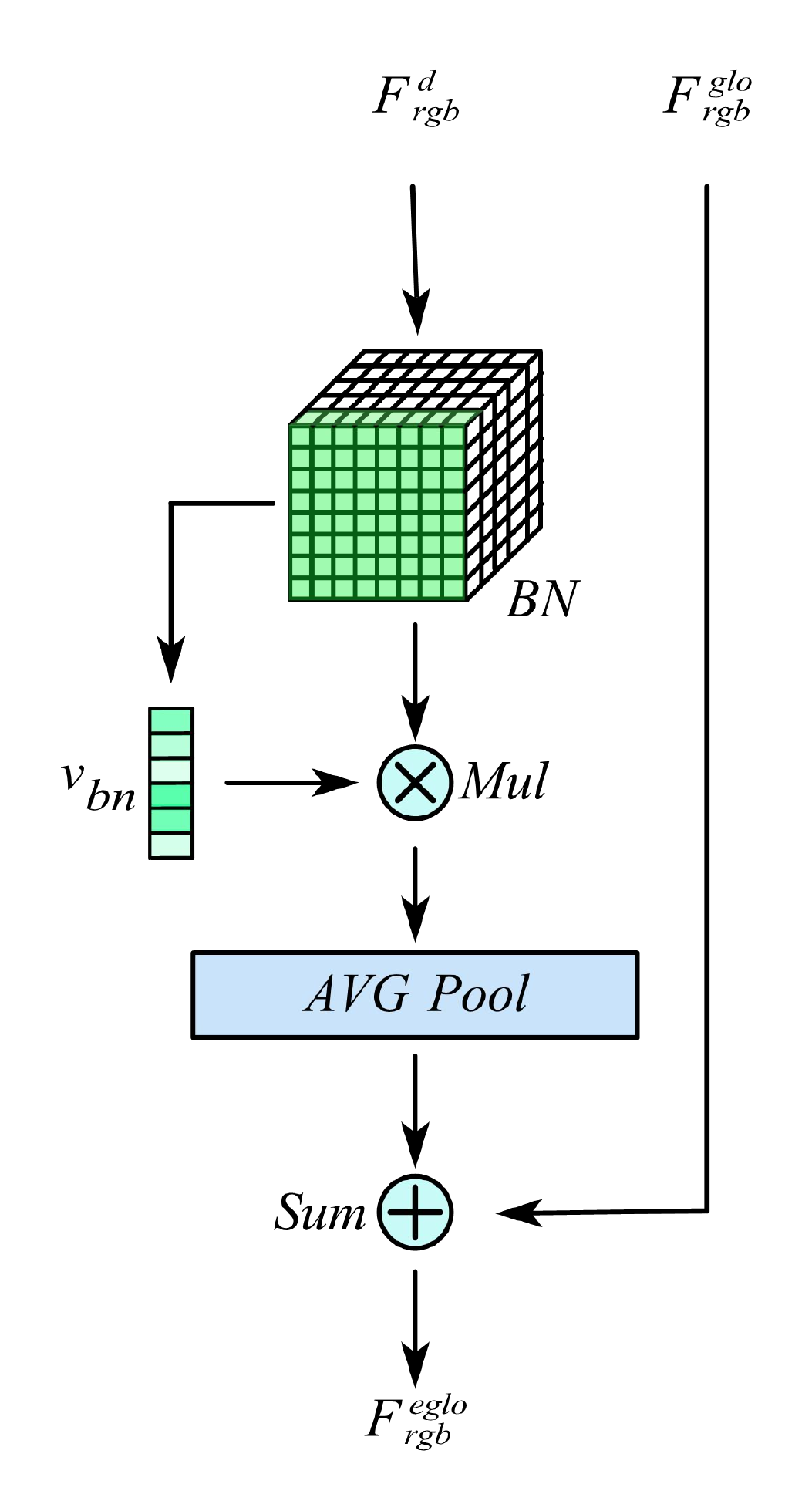}
		\vspace{-.05in}
		\caption{Global feature enhancement module}
		\label{fig5}
	\end{figure}

It is no doubt that global features play an important role in person re-identification, and a BN-based global feature enhancement module is designed to further improve the discrimination of different pedestrian global representations under cross-modality. As shown in  Fig.~\ref{fig5}, taking the visible mode as an example.

Firstly, the unified spatial output features $F_{rgb}^d$ are normalized through the batch-normalization layer (BN) At the same time, the vector ${v_{bn}}$ is calculated according to the BN weights. Normalize the features $F_{rgb}^d$ using BN and multiply them with ${v_{bn}}$ to obtain a new feature and use this feature to obtain a new weight matrix by adaptive averaging pooling layer. To obtain the final $F_{rgb}^{eglo}$, we linearly sum $F_{rgb}^{glo}$ and this feature.

Then, we derive the calculation process from the perspective of the formula. We first define the BN layer calculation, as in Eq.~\ref{eq:4}:
\begin{equation}
	\small
	\label{eq:4}
	BN(x) = \gamma \frac{{x - {\mu _b}}}{{\sqrt {\sigma _b^2 + \varepsilon } }} + \beta	
\end{equation}
where $\gamma$ is scale factor and $\beta$ is translation factor, ${\mu _b}$ and $\sigma _b^2$ is mean and variance of one batch, respectively. $\varepsilon$ is a hyper parameter.

We define the weight vector ${v_{bn}} = {\lambda _j}/\sum\limits_{j = 0} {{\lambda _j}}$,where ${\lambda _j}$is the weight factor in the BN layer.

Finally, The enhanced global features are calculated as in Eq.~\ref{eq:5}:
\begin{equation}
	\small
	\label{eq:5}
	F_{rgb}^{eglo} = F_{rgb}^{glo} + Avg({v_{bn}} \bullet BN(F_{rgb}^d))
\end{equation}
where $Avg$ is adaptive average pooling calculation process. 
\subsection{The Multi-granularity Fusion Loss}
The multi-granularity fusion loss proposed in this paper is described in detail below. We use two-stream networks to obtain global and local features, and then use them to compute classification loss and triple loss. Classification loss is widely used in the Reid task to calculate cross-entropy loss mainly by pedestrian identity labels, which is referred to as id loss in this paper. 

Firstly, we define the global and local classification losses as $L_{id}^g$ and $L_{id}^{lv}$ , respectively, as Eq.~\ref{eq:6} and Eq.~\ref{eq:7}:
\begin{equation}
	\small
	\label{eq:6}
	L_{id}^g = \sum\limits_{i = 1}^N { - {q_i}} \log (p_i^g)
\end{equation}
\begin{equation}
	\small
	\label{eq:7}
L_{id}^{lv} = \sum\limits_{j = 2}^S {\sum\limits_{i = 1}^N { - {q_i}} \log (p_i^j)} {\rm{}}
\end{equation}
where $N$ is the total number of categories in the training dataset, ${q_i}$ is the sample true probability distribution , $S$ is the number of horizontal slices, $p_i^j$ and $p_i^g$ are the predicted probability distributions.

For each of $P$ that randomly selected person identities, $K$ visible images and $K$ thermal images are randomly sampled, totally are $2 \times P \times K$ images, We define the heterogeneous center-based triad loss as Eq.~\ref{eq:8}
\begin{equation}
	\small	
	\label{eq:8}
	\begin{aligned}	
	L_{Tri}^g = \sum\limits_{i = 1}^P {{{[{m_g} + {{\left\| {fc_v^i - fc_t^i} \right\|}_2} - \mathop {\min }\limits_{k \in \{ v,t\} } {{\left\| {fc_v^i - fc_k^j} \right\|}_2}]}_ + }}&\\
	 + \sum\limits_{i = 1}^P {{{[{m_g} + {{\left\| {fc_t^i - fc_v^i} \right\|}_2} - \mathop {\min }\limits_{k \in \{ v,t\} } {{\left\| {fc_t^i - fc_k^j} \right\|}_2}]}_ + }}& 
\end{aligned}
\end{equation}

where ${m_g}$ is a hyper parameter.$fc_v^i = \frac{1}{K}\sum\nolimits_{j = 1}^K {f_{v,j}^i}$ and $fc_t^i = \frac{1}{K}\sum\nolimits_{j = 1}^K {f_{t,j}^i}$ are the features centers of different modality. $fc_v^j$ and $fc_t^j$ represent pedestrian features in different modality respectively.

Considering the auxiliary role of local features to the performance of the algorithm, the cross-modal local feature alignment loss $L_{PA}^{vt}$ is designed in this paper and defined as Eq.~\ref{eq:9}:
\begin{equation}
	\small
	\label{eq:9}
	\begin{aligned}	
	L_{PA}^{vt} = \sum\limits_{i = 1}^P {\sum\limits_{a = 1}^{2K} {\sum\limits_{j = 2}^H {[{m_l} + \mathop {\max }\limits_{k \in \{ v,t\} } {{\left\| {f_{i.j}^{ka} - f_{i,j}^{kp}} \right\|}_2}} } }&\\
	{\rm{ }} - \mathop {\min }\limits_{k \in \{ v,t\} } {\left\| {f_{i,j}^{ka} - f_{i,j}^{kn}} \right\|_2}]&	
\end{aligned}
\end{equation}
where ${m_l}$ is a hyper parameter, $f_{i.j}^{ka}$ is the infrared/visible modal local features, $f_{i,j}^{kp}$ is the visible/infrared mode with the most distant positive sample, $f_{i,j}^{kn}$is the visible/infrared mode  distance nearest negative sample and $k$ represents different modes.
In this paper, the total loss ${L_{total}}$ is defined by the Eq.~\ref{eq:10}:
\begin{equation}
	\small
	\label{eq:10}
	{L_{total}} = \overbrace {L_{id}^g + L_{id}^{eg}}^{Global{\rm{ }}id} + \overbrace {L_{Tri}^g + L_{Tri}^{eg}}^{Global{\rm{ }}Tri} + \overbrace {L_{id}^{lv} + L_{id}^{lt}}^{Local{\rm{ }}id} + \overbrace {L_{PA}^{vt}}^{Local{\rm{ }}Tri}
\end{equation}

\section{Experiment}
\label{section4}
%-------------------------------------------------------------------------
\subsection{Experimental settings}
\textbf{Datasets.} 
SYSU-MM01 is the first cross-modal pedestrian dataset built by Wu et al. All data are collected from 4 RGB cameras and 2 IR cameras, including a total of 491 different pedestrians. 395 people are included in the training set, with a total of 19,659 visible images and 12,792 IR images, and 96 people are included in the test set, including full-scene search and indoor scene search. For indoor search only search in cameras1,2,3and6, while the full scene uses all images.

RegDB\textsuperscript{\cite{nguyen2017person}} contains 412 individuals, each with 10 images from the visible camera and 10 images from the infrared camera.

\textbf{Evaluation metrics.} 
The CMC (Cumulative Matching Characteristics), the mINP (mean inverse negative penalty) and mAP (mean Average Precision) are used to evaluate the retrieval performance. For CMC, we report the rank-1, rank-10, and rank-20 precision.

Official method evaluation was performed on the SYSU-MM01dataset, based on 10 replicate random splits of the query set dataset and the mean of the group to be tested. RegDB results were based on 10 replicate random splits of the training and test sets and the mean was calculated.

\textbf{Implementation Details.} 
We implemented our model with Pytorch. The training input images are first padded with 10 and randomly cropped to 288×144,random horizontal flipping is further imposed as data augmentation. The train-batch size is set to 120.  We set the ImageNet pre-trained CNN part as 0.01 and the classifier as 0.1, optimized them via SGD. The learning rate decreases by a factor of 10 every 10 epochs.

\textbf{Re-ranking.} 
In recent studies\textsuperscript{\cite{zhang2017alignedreid,basaran2020efficient,sarfraz2018pose}}, the Re-ranking algorithm ,which is an algorithm for reordering Retrieval Results, contributes significantly to person Re-Identification, so that we used ECN\textsuperscript{\cite{sarfraz2018pose}}(Expanded Cross Neighborhood) as a post-processing algorithm in the inference phase. We used the relevant parameter sources in 35.

% Please add the following required packages to your document preamble:
% \usepackage{multirow}
\begin{table*}[]
	\centering
	\small 
	\caption{Comparison with the state-of-the-art methods on the SYSU-MM01 dataset.}
	\label{tab sysu}
	
	\begin{tabular}{c|c|ccccc|ccccc}
		\hline
		\multirow{2}{*}{Methods} & \multirow{2}{*}{Time} & \multicolumn{5}{c|}{All search}                                                                                                       & \multicolumn{5}{c}{Indoor search}                                                                                                     \\ \cline{3-12} 
		&                       & \multicolumn{1}{c|}{rank-1} & \multicolumn{1}{c|}{rank-10} & \multicolumn{1}{c|}{rank-20} & \multicolumn{1}{c|}{mAP} & mINP           & \multicolumn{1}{c|}{rank-1} & \multicolumn{1}{c|}{rank-10} & \multicolumn{1}{c|}{rank-20} & \multicolumn{1}{c|}{mAP} & mINP           \\ \hline
		HOG~\cite{dalal2005histograms}                      & 2005                  & 2.76                        & 18.25                        & 31.91                        & 4.24                     & -              & 3.22                        & 24.68                        & 44.52                        & 7.25                     & -              \\ \hline
		LOMO~\cite{liao2015person}                     & 2015                  & 3.64                        & 23.18                        & 37.28                        & 4.53                     & -              & 5.75                        & 34.35                        & 57.90                        & 10.19                    & -              \\ \hline
		HCML~\cite{ye2018hierarchical}                     & 2018                  & 14.32                       & 53.16                        & 69.17                        & 16.16                    & -              & 20.58                       & 63.38                        & 85.79                        & 26.92                    & -              \\
		DCTR~\cite{ye2018visible}                     & 2018                  & 17.01                       & 55.43                        & 71.96                        & 19.66                    & -              & -                           & -                            &                              & -                        & -              \\
		cmGAN~\cite{dai2018cross}                    & 2018                  & 26.97                       & 67.51                        & 80.56                        & 31.49                    & -              & 31.63                       & 77.23                        & 89.62                        & 42.46                    & -              \\ \hline
		HSME~\cite{hao2019hsme}                     & 2019                  & 20.68                       & 62.74                        & 77.95                        & 23.12                    & -              & -                           & -                            & -                            & -                        & -              \\
		IPVT+MSR~\cite{kang2019person}                 & 2019                  & 23.20                       & 51.20                        & 61.70                        & 22.50                    & -              & -                           & -                            & -                            & -                        & -              \\
		AlignGAN~\cite{wang2019rgb}                 & 2019                  & 51.50                       & 89.40                        & 95.70                        & 33.90                    & -              & 57.10                       & 92.70                        & 97.40                        & 45.30                    & -              \\ \hline
		CoSiGAN~\cite{zhong2020visible}                  & 2020                  & 35.55                       & 81.54                        & 90.43                        & 38.33                    & -              & -                           & -                            & -                            & -                        & -              \\
		EDFL~\cite{liu2020enhancing}                     & 2020                  & 36.94                       & 84.52                        & 93.22                        & 40.77                    &                & -                           & -                            & -                            & -                        & -              \\
		HPILN~\cite{zhao2019hpiln}                    & 2020                  & 41.36                       & 84.78                        & 94.51                        & 42.95                    & -              & 45.77                       & 91.82                        & 98.46                        & 56.52                    & -              \\
		JSIA~\cite{wang2020cross}                     & 2020                  & 45.10                       & 85.70                        & 93.80                        & 29.50                    & -              & 52.70                       & 91.10                        & 96.40                        & 42.70                    & -              \\
		CML~\cite{ling2020class}                      & 2020                  & 56.27                       & 94.08                        & 98.12                        & 43.39                    & -              & 60.42                       & 95.88                        & 99.5                         & 53.52                    & -              \\
		HC~\cite{zhu2020hetero}                       & 2020                  & 59.96                       & 91.50                        & 96.82                        & 54.95                    & -              & 59.74                       & 92.07                        & 96.22                        & 64.91                    & -              \\
		HcTri~\cite{liu2020parameter}                    & 2020                  & 61.68                       & 93.10                        & 97.17                        & 57.51                    & 39.54          & 63.41                       & 91.69                        & 95.28                        & 68.10                    & 64.26          \\
		LZM~\cite{basaran2020efficient}                       & 2020                  & 63.05                       & 93.62                        & 96.30                        & 67.13                    & -              & 69.06                       & 96.30                        & 97.16                        & 76.95                    & -              \\ \hline
		TS-GAN~\cite{zhang2021rgb}                    & 2021                  & 55.90                       & 91.20                        & 96.60                        & 39.70                    & -              & 59.30                       & 91.80                        & 97.90                        & 50.90                    & -              \\
		AGW~\cite{xia2021visible}                       & 2021                  & 56.52                       & 90.26                        & 95.59                        & 57.47                    & 38.75          & 68.72                       & 94.61                        & 97.42                        & 75.11                    & 64.22          \\
		DGTL~\cite{liu2021strong}                     & 2021                  & 57.34                       & -                            & -                            & 55.13                    & -              & 63.11                       & -                            & 69.2                         & -                        & -              \\
		MPANet~\cite{wu2021discover}                    & 2021                  & 70.58                       & 96.21                        & 98.80                        & 68.24                    & -              & 76.74                       & 98.21                        & 99.57                        & 80.95                    & -              \\ \hline
		CM-EMD~\cite{ling2022cross}                   & 2022                  & 73.39                       & \textbf{96.24}               & \textbf{98.82}               & 68.56                    & -              & 80.53                       & \textbf{98.31}               & 99.91                        & 82.71                    & -              \\ \hline
		\textbf{CM-GE}                    & -                     & 58.59                       & 87.14                        & 92.64                        & 54.95                    & 39.00          & 65.84                       & 93.22                        & 96.95                        & 67.31                    & 61.99          \\
		\textbf{CM-LSP-GE}                   & -                     & 60.91                       & 89.03                        & 94.43                        & 57.12                    & 41.39          & 69.38                       & 95.14                        & 98.25                        & 70.3                     & 65.14          \\
		\textbf{CM-LSP-GE-RK}             & -                     & \textbf{76.28}              & 94.38                        & 97.08                        & \textbf{76.52}           & \textbf{42.03} & \textbf{82.31}              & 98.12                        & \textbf{99.91}               & \textbf{85.16}           & \textbf{66.51} \\ \hline
	\end{tabular}
\end{table*}

\subsection{Comparison with The State-of-The-Art Algorithms}

To demonstrate the superiority of our method, we compared it against the state-of-the-art approaches on SYSU-MM01 and RegDB. These methods include:traditional characterization:HOG~\cite{dalal2005histograms},LOMO~\cite{liao2015person},deep learning:HCML~\cite{ye2018hierarchical},LZM~\cite{basaran2020efficient},IPVT+MSR~\cite{kang2019person}.

Metric Learning Methods: HC~\cite{zhu2020hetero},DCTR~\cite{ye2018visible},CML~\cite{ling2020class}
,EDFL~\cite{liu2020enhancing},HSME~\cite{hao2019hsme},HPILN~\cite{zhao2019hpiln},DGTL~\cite{liu2021strong},HcTri~\cite{liu2020parameter},CM-EMD~\cite{ling2022cross},MPANet~\cite{wu2021discover}.

Image generation methods:cmGAN~\cite{dai2018cross},AlignGAN~\cite{wang2019rgb},
JSIA~\cite{wang2020cross}, CoSiGAN~\cite{zhong2020visible},TS-GAN~\cite{zhang2021rgb}.
The experimental results respectively correspond to Tabel~\ref{tab regdb} and Tabel~\ref{tab sysu}.
% Please add the following required packages to your document preamble:
% \usepackage{multirow}

\textbf{Results on SYSU-MM01}.
The results are shown in Tabel~\ref{tab sysu}. Our method respectively obtains 76.28\% and 76.52\% of Rank-1 and mAP accuracy in the full scene search, and obtain 82.31\% and 85.16\% of Rank-1 and mAP accuracy in the indoor search. The test results are better than the latest methods.
\begin{table}[]
	\centering
	\small 
	\caption{Comparison with the state-of-the-art methods on the RegDB dataset.}
	\label{tab regdb}
	\begin{tabular}{c|c|cc|cc}
		\hline
		\multirow{2}{*}{Methods} & \multirow{2}{*}{Time} & \multicolumn{2}{c|}{Visible-Thermal}         & \multicolumn{2}{c}{Thermal- Visible}         \\ \cline{3-6} 
		&                       & \multicolumn{1}{c|}{rank-1} & mAP            & \multicolumn{1}{c|}{rank-1} & mAP            \\ \hline
		HCML~\cite{ye2018hierarchical}                     & 2018                  & 24.44                       & 20.08          & 21.70                       & 22.24          \\
		DCTR~\cite{liu2021strong}                     & 2018                  & 33.47                       & 31.83          & 32.72                       & 31.10          \\ \hline
		HSME~\cite{hao2019hsme}                     & 2019                  & 50.85                       & 47.00          & 50.15                       & 46.16          \\
		AlignGAN~\cite{wang2019rgb}                 & 2019                  & 57.9                        & 53.6           & 56.3                        & 53.4           \\ \hline
		CoSiGAN~\cite{zhong2020visible}                  & 2020                  & 47.18                       & 46.16          & -                           & -              \\
		EDFL~\cite{liu2020enhancing}                      & 2020                  & 52.58                       & 52.98          & 51.89                       & 52.13          \\
		CML~\cite{ling2020class}                      & 2020                  & -                           & -              & 59.81                       & 60.86          \\
		HcTri~\cite{liu2020parameter}                    & 2020                  & 91.05                       & 83.28          & 89.3                        & 81.46          \\
		LZM~\cite{basaran2020efficient}                      & 2020                  & 60.58                       & 63.36          & 57.17                       & 57.56          \\ \hline
		AGW~\cite{xia2021visible}                       & 2021                  & 78.3                        & 70.37          & 75.22                       & 67.28          \\
		DGTL~\cite{liu2021strong}                     & 2021                  & 83.92                       & 73.78          & 81.59                       & 71.65          \\
		MPANet~\cite{wu2021discover}                   & 2021                  & 83.7                        & 80.9           & 82.8                        & 80.7           \\ \hline
		CM-EMD~\cite{ling2022cross}                   & 2022                  & \textbf{94.37}              & 88.23          & 92.77                       & 86.85          \\ \hline
		\textbf{CM-GE}           & -                     & 92.32                       & 85.35          & 90.16                       & 83.43          \\
		\textbf{CM-LSP-GE}       & -                     & 94.13                       & \textbf{88.86} & \textbf{93.16}              & \textbf{87.26} \\ \hline
	\end{tabular}
\end{table}

\textbf{Results on RegDB}.
We use the method of this paper on the RegDB dataset to compare with the current state-of-the-art methods, and the results are shown in Tabel~\ref{tab regdb}. In Visible-Thermal test results, our method achieves Rank-1 accuracy of 94.13\% and mAP accuracy of 88.86\%, and in Thermal-Visible test our method obtains Rank-1 accuracy of 93.16\% and mAP accuracy of 87.26\%. In VT test, the results are basically the same as the latest literature CM-EMD~\cite{ling2022cross} results were essentially unchanged and in the TV test, 0.39\% improvement was achieved on Rank-1 and 0.41\% improvement on mAP. Compared with HcTri~\cite{liu2020parameter}, our method has better performance in all aspects.

\subsection{Evaluation.}
\begin{figure*}[!t]
	\centering
	\includegraphics[width=0.95\linewidth]{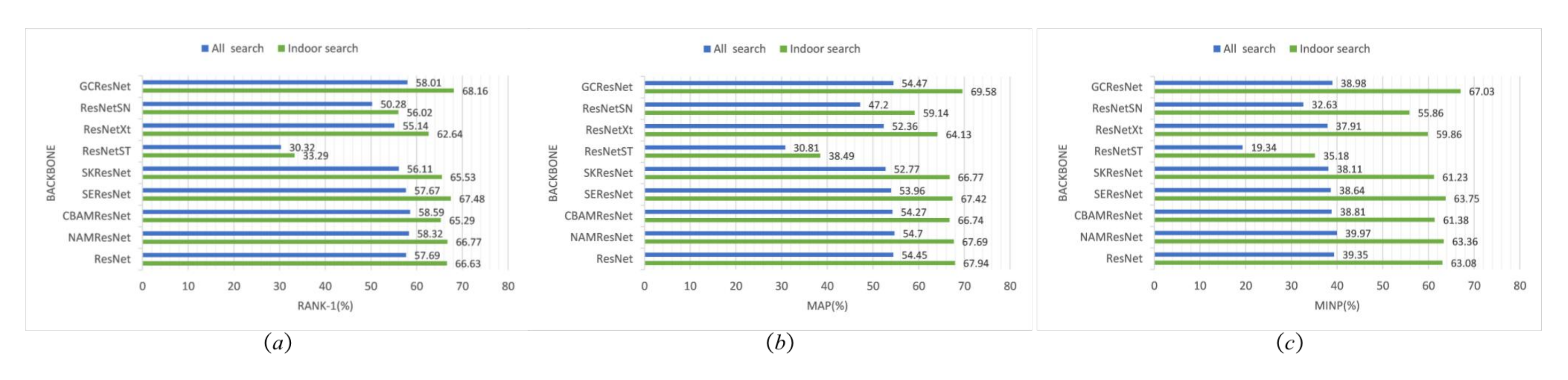}
	\caption{(a) Impacts of Backbone in terms of Rank-1,(b) Impacts of Backbone in terms of mAP, (c) Impacts of Backbone in terms of mINP}
	\label{fig6}
\end{figure*}

\textbf{Effectiveness of attention mechanism}.
In order to study the effect of different attention methods on re-identification accuracy, the following eight different methods are selected for experimental study. such as SEResNet based on channel attention, CBAMResNet based on convolutional block attention, GCResNet based on full-text information, ResNetST based on scattered attention, NAMResNet based on normalized attention, SKResNet based on selective kernel, ResNetXT based on design aggregated residual block, ResNetSN based on switchable normalized layer. As shown in Fig.~\ref{fig6}, the experimental results prove the improved backbone network based on GC attention mechanism has the best performance.

\begin{figure*}[!t]
	\centering
	\includegraphics[width=0.95\linewidth]{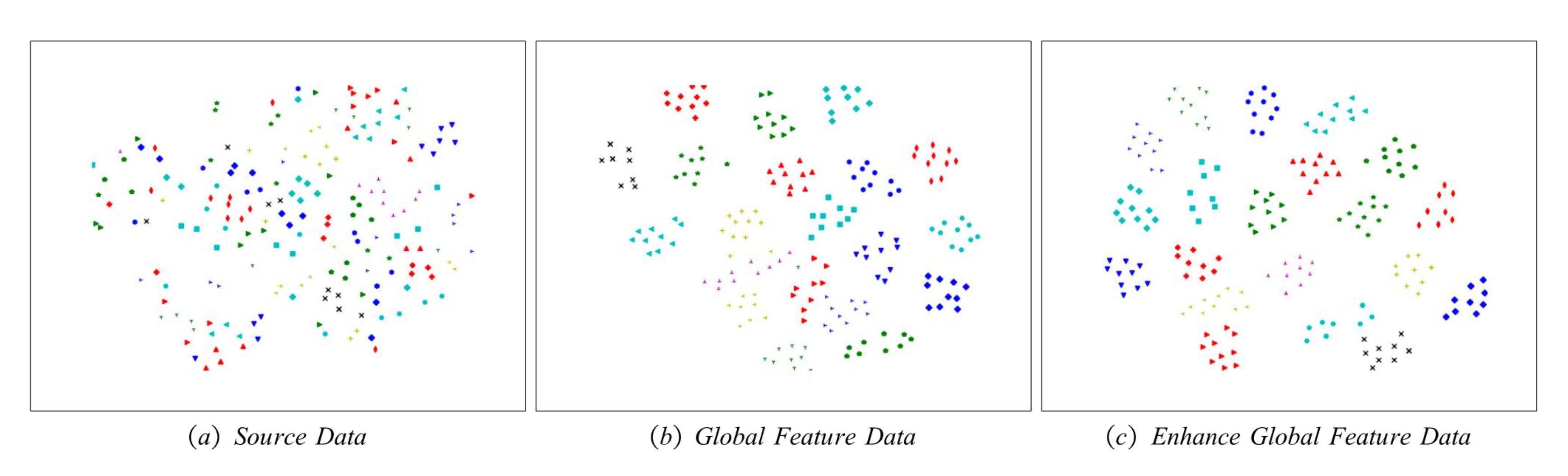}
	\caption{(a) Impacts of Backbone in terms of Rank-1,(b) Impacts of Backbone in terms of mAP, (c) Impacts of Backbone in terms of mINP}
	\label{fig7}
\end{figure*}

\textbf{Effectiveness of GE}.
To verify the effectiveness of the GE module designed in this paper, we select global features and enhanced global features in the network for visualization and analysis which are shown in Figure 7. We randomly select 20 different pedestrians from the SYSU-MM01 query dataset, and take 10 images of each person separately for visualization and analysis. In Fig.~\ref{fig7} (a) the image is mapped to a two-dimensional plane using PCA, (b) shows the global features mapped onto the plane visualization and (c) shows the enhanced global features. Comparing (b) and (c), it can be found that the distance between different classes is farther after global feature enhancement. 

\textbf{Effectiveness of CM-LSP}.
In order to verify the effectiveness of the proposed method for cross-modal local feature alignment based on the shortest path the cutting quantity experiment on SYSU-MM01 is firstly carried out. From the result in Tabel~\ref{tab3}.it can be found that the best result is obtained by dividing the body image into three equal parts.

We chose the number of parts in Tabel~\ref{tab3}. as 2 for the validation dimension experiment and selectively verified that the dimensions 128, 256, 516 and 1024. As shown in Tabel~\ref{tab4}, we found that the 1024 dims features have the best performance from the combined results.

\begin{table}[]
	\centering	
	\normalsize 
	\caption{Comparison with Part quantity in SYSU-MM01 do indoor search.}
	\label{tab3}
	\begin{tabular}{c|ccc}
		\hline	
		Parts & \multicolumn{1}{c|}{Rank-1} & \multicolumn{1}{c|}{mAP} & mINP   \\ \hline
		None  & 65.84                       & 67.31                    & 61.99  \\
		2     & 67.26                       & 68.08                    & 62.52  \\
		3     & \textbf{69.38}                       & \textbf{70.30}                    & \textbf{65.14}  \\
		4     & 68.55                       & 69.47                    & 64.09  \\
		5     & 66.66                       & 68.22                    & 62.42  \\
		6     & 67.82                       & 68.80                    & 63.28 \\
		7     & 67.76                       & 68.56                    & 62.62  \\
		8     & 67.17                       & 68.21                    & 62.90  \\
		9     & 66.02                       & 67.52                    & 62.50  \\ 
	\hline
	\end{tabular}
\end{table}

\begin{table}[]
	\centering	
	\normalsize 
	\caption{Comparison with two Part and differ Dim in SYSU-MM01 do indoor search.}
	\label{tab4}
	\begin{tabular}{l|lllll}
		\hline
		Dim  & \multicolumn{1}{l|}{rank-1} & \multicolumn{1}{l|}{rank-10} & \multicolumn{1}{l|}{rank-20} & \multicolumn{1}{l|}{mAP} & mINP  \\ \hline
		128  & \textbf{67.26}                       & 93.79                        & 97.15                        & 68.08                    & 62.52 \\
		256  & 65.25                       & \textbf{93.96}                        & 97.21                        & 66.65                    & 60.92 \\
		512  & 65.58                       & 93.51                        & \textbf{97.40}                        & 66.88                    & 61.04 \\
		1024 & 67.13                       & 93.72                        & 97.35                        & \textbf{68.72}                    & \textbf{62.78}  \\ \hline
	
	\end{tabular}
\end{table}

Experimental results demonstrated that joint learning using global and local features is superior to global features alone, and with the help of local features, the network pays more attention to the similarity of pedestrian body parts in images rather than overfitting background information.

\section{Conclusion}
\label{sec:conclusion}
Considering the important role of feature fusion of global and local features for the overall re-identification task accuracy improvement,we proposed a novel Visible-Thermal person re-Identification network in this paper, called Cross-modal Local Shortest path and Batch Normalized Global Enhancement (CM-LSP-GE), which can mitigate cross-modal differences and resolve occlusion problems in images. Moreover, we introduce four different methods to facilitate the CM-LSP-GE proposed in this paper, which are the attention-based two-stream network, cross-modal local feature alignment based on the shortest path (CM-LSP),batch normalized global feature enhancement (GE) and the multi-granularity fusion loss.CM-LSP can ensure that the model learns pedestrian details in the image without overfitting the background at a fine-grained level, while GE can enhance the global feature representation of the image at a coarse-grained level. Loss function guides network learning from global and local levels. Eventually, the accuracy is effectively improved, and our method achieves better results on RegDB and SYSU-MM01.

	%------------------------------------------------------------------------
	{\small
		\bibliographystyle{IEEEtran}
		\bibliography{ref.bib}

% Generated by IEEEtran.bst, version: 1.14 (2015/08/26)
\begin{thebibliography}{10}
\providecommand{\url}[1]{#1}
\csname url@samestyle\endcsname
\providecommand{\newblock}{\relax}
\providecommand{\bibinfo}[2]{#2}
\providecommand{\BIBentrySTDinterwordspacing}{\spaceskip=0pt\relax}
\providecommand{\BIBentryALTinterwordstretchfactor}{4}
\providecommand{\BIBentryALTinterwordspacing}{\spaceskip=\fontdimen2\font plus
\BIBentryALTinterwordstretchfactor\fontdimen3\font minus
  \fontdimen4\font\relax}
\providecommand{\BIBforeignlanguage}[2]{{%
\expandafter\ifx\csname l@#1\endcsname\relax
\typeout{** WARNING: IEEEtran.bst: No hyphenation pattern has been}%
\typeout{** loaded for the language `#1'. Using the pattern for}%
\typeout{** the default language instead.}%
\else
\language=\csname l@#1\endcsname
\fi
#2}}
\providecommand{\BIBdecl}{\relax}
\BIBdecl

\bibitem{ye2021deep}
M.~Ye, J.~Shen, G.~Lin, T.~Xiang, L.~Shao, and S.~C. Hoi, ``Deep learning for
  person re-identification: A survey and outlook,'' \emph{IEEE transactions on
  pattern analysis and machine intelligence}, vol.~44, no.~6, pp. 2872--2893,
  2021.

\bibitem{wu2017rgb}
A.~Wu, W.-S. Zheng, H.-X. Yu, S.~Gong, and J.~Lai, ``Rgb-infrared
  cross-modality person re-identification,'' in \emph{Proceedings of the IEEE
  international conference on computer vision}, 2017, pp. 5380--5389.

\bibitem{dalal2005histograms}
N.~Dalal and B.~Triggs, ``Histograms of oriented gradients for human
  detection,'' in \emph{2005 IEEE computer society conference on computer
  vision and pattern recognition (CVPR'05)}, vol.~1.\hskip 1em plus 0.5em minus
  0.4em\relax Ieee, 2005, pp. 886--893.

\bibitem{liao2015person}
S.~Liao, Y.~Hu, X.~Zhu, and S.~Z. Li, ``Person re-identification by local
  maximal occurrence representation and metric learning,'' in \emph{Proceedings
  of the IEEE conference on computer vision and pattern recognition}, 2015, pp.
  2197--2206.

\bibitem{kniaz2018thermalgan}
V.~V. Kniaz, V.~A. Knyaz, J.~Hladuvka, W.~G. Kropatsch, and V.~Mizginov,
  ``Thermalgan: Multimodal color-to-thermal image translation for person
  re-identification in multispectral dataset,'' in \emph{Proceedings of the
  European Conference on Computer Vision (ECCV) Workshops}, 2018, pp. 0--0.

\bibitem{zhang2021rgb}
Z.~Zhang, S.~Jiang, C.~Huang, Y.~Li, and R.~Y. Da~Xu, ``Rgb-ir cross-modality
  person reid based on teacher-student gan model,'' \emph{Pattern Recognition
  Letters}, vol. 150, pp. 155--161, 2021.

\bibitem{xia2021visible}
D.~Xia, H.~Liu, L.~Xu, and L.~Wang, ``Visible-infrared person re-identification
  with data augmentation via cycle-consistent adversarial network,''
  \emph{Neurocomputing}, vol. 443, pp. 35--46, 2021.

\bibitem{kang2019person}
J.~K. Kang, T.~M. Hoang, and K.~R. Park, ``Person re-identification between
  visible and thermal camera images based on deep residual cnn using single
  input,'' \emph{IEEE Access}, vol.~7, pp. 57\,972--57\,984, 2019.

\bibitem{fan2020cross}
X.~Fan, H.~Luo, C.~Zhang, and W.~Jiang, ``Cross-spectrum dual-subspace pairing
  for rgb-infrared cross-modality person re-identification,'' \emph{arXiv
  preprint arXiv:2003.00213}, 2020.

\bibitem{ye2018hierarchical}
M.~Ye, X.~Lan, J.~Li, and P.~Yuen, ``Hierarchical discriminative learning for
  visible thermal person re-identification,'' in \emph{Proceedings of the AAAI
  Conference on Artificial Intelligence}, vol.~32, no.~1, 2018.

\bibitem{jiang2020cross}
J.~Jiang, K.~Jin, M.~Qi, Q.~Wang, J.~Wu, and C.~Chen, ``A cross-modal
  multi-granularity attention network for rgb-ir person re-identification,''
  \emph{Neurocomputing}, vol. 406, pp. 59--67, 2020.

\bibitem{ran2021improving}
L.~Ran, Y.~Hong, S.~Zhang, Y.~Yang, and Y.~Zhang, ``Improving visible-thermal
  reid with structural common space embedding and part models,'' \emph{Pattern
  Recognition Letters}, vol. 142, pp. 25--31, 2021.

\bibitem{basaran2020efficient}
E.~Basaran, M.~G{\"o}kmen, and M.~E. Kamasak, ``An efficient framework for
  visible--infrared cross modality person re-identification,'' \emph{Signal
  Processing: Image Communication}, vol.~87, p. 115933, 2020.

\bibitem{hu2018squeeze}
J.~Hu, L.~Shen, and G.~Sun, ``Squeeze-and-excitation networks,'' in
  \emph{Proceedings of the IEEE conference on computer vision and pattern
  recognition}, 2018, pp. 7132--7141.

\bibitem{woo2018cbam}
S.~Woo, J.~Park, J.-Y. Lee, and I.~S. Kweon, ``Cbam: Convolutional block
  attention module,'' in \emph{Proceedings of the European conference on
  computer vision (ECCV)}, 2018, pp. 3--19.

\bibitem{cao2019gcnet}
Y.~Cao, J.~Xu, S.~Lin, F.~Wei, and H.~Hu, ``Gcnet: Non-local networks meet
  squeeze-excitation networks and beyond,'' in \emph{Proceedings of the
  IEEE/CVF International Conference on Computer Vision Workshops}, 2019, pp.
  0--0.

\bibitem{li2019selective}
X.~Li, W.~Wang, X.~Hu, and J.~Yang, ``Selective kernel networks,'' in
  \emph{Proceedings of the IEEE/CVF Conference on Computer Vision and Pattern
  Recognition}, 2019, pp. 510--519.

\bibitem{zhang2020resnest}
H.~Zhang, C.~Wu, Z.~Zhang, Y.~Zhu, H.~Lin, Z.~Zhang, Y.~Sun, T.~He, J.~Mueller,
  R.~Manmatha \emph{et~al.}, ``Resnest: Split-attention networks,'' \emph{arXiv
  preprint arXiv:2004.08955}, 2020.

\bibitem{liu2021nam}
Y.~Liu, Z.~Shao, Y.~Teng, and N.~Hoffmann, ``Nam: Normalization-based attention
  module,'' \emph{arXiv preprint arXiv:2111.12419}, 2021.

\bibitem{xie2017aggregated}
S.~Xie, R.~Girshick, P.~Doll{\'a}r, Z.~Tu, and K.~He, ``Aggregated residual
  transformations for deep neural networks,'' in \emph{Proceedings of the IEEE
  conference on computer vision and pattern recognition}, 2017, pp. 1492--1500.

\bibitem{luo2018differentiable}
P.~Luo, J.~Ren, Z.~Peng, R.~Zhang, and J.~Li, ``Differentiable
  learning-to-normalize via switchable normalization,'' \emph{arXiv preprint
  arXiv:1806.10779}, 2018.

\bibitem{ye2020bi}
H.~Ye, H.~Liu, F.~Meng, and X.~Li, ``Bi-directional exponential angular triplet
  loss for rgb-infrared person re-identification,'' \emph{IEEE Transactions on
  Image Processing}, vol.~30, pp. 1583--1595, 2020.

\bibitem{ye2018visible}
M.~Ye, Z.~Wang, X.~Lan, and P.~C. Yuen, ``Visible thermal person
  re-identification via dual-constrained top-ranking.'' in \emph{IJCAI},
  vol.~1, 2018, p.~2.

\bibitem{zhu2020hetero}
Y.~Zhu, Z.~Yang, L.~Wang, S.~Zhao, X.~Hu, and D.~Tao, ``Hetero-center loss for
  cross-modality person re-identification,'' \emph{Neurocomputing}, vol. 386,
  pp. 97--109, 2020.

\bibitem{ling2020class}
Y.~Ling, Z.~Zhong, Z.~Luo, P.~Rota, S.~Li, and N.~Sebe, ``Class-aware modality
  mix and center-guided metric learning for visible-thermal person
  re-identification,'' in \emph{Proceedings of the 28th ACM International
  Conference on Multimedia}, 2020, pp. 889--897.

\bibitem{liu2020enhancing}
H.~Liu, J.~Cheng, W.~Wang, Y.~Su, and H.~Bai, ``Enhancing the discriminative
  feature learning for visible-thermal cross-modality person
  re-identification,'' \emph{Neurocomputing}, vol. 398, pp. 11--19, 2020.

\bibitem{hao2019hsme}
Y.~Hao, N.~Wang, J.~Li, and X.~Gao, ``Hsme: Hypersphere manifold embedding for
  visible thermal person re-identification,'' in \emph{Proceedings of the AAAI
  conference on artificial intelligence}, vol.~33, no.~01, 2019, pp.
  8385--8392.

\bibitem{zhao2019hpiln}
Y.-B. Zhao, J.-W. Lin, Q.~Xuan, and X.~Xi, ``Hpiln: a feature learning
  framework for cross-modality person re-identification,'' \emph{IET Image
  Processing}, vol.~13, no.~14, pp. 2897--2904, 2019.

\bibitem{liu2021strong}
H.~Liu, Y.~Chai, X.~Tan, D.~Li, and X.~Zhou, ``Strong but simple baseline with
  dual-granularity triplet loss for visible-thermal person re-identification,''
  \emph{IEEE Signal Processing Letters}, vol.~28, pp. 653--657, 2021.

\bibitem{liu2020parameter}
H.~Liu, X.~Tan, and X.~Zhou, ``Parameter sharing exploration and hetero-center
  triplet loss for visible-thermal person re-identification,'' \emph{IEEE
  Transactions on Multimedia}, vol.~23, pp. 4414--4425, 2020.

\bibitem{ling2022cross}
Y.~Ling, Z.~Zhong, D.~Cao, Z.~Luo, Y.~Lin, S.~Li, and N.~Sebe, ``Cross-modality
  earth mover's distance for visible thermal person re-identification,''
  \emph{arXiv preprint arXiv:2203.01675}, 2022.

\bibitem{wu2021discover}
Q.~Wu, P.~Dai, J.~Chen, C.-W. Lin, Y.~Wu, F.~Huang, B.~Zhong, and R.~Ji,
  ``Discover cross-modality nuances for visible-infrared person
  re-identification,'' in \emph{Proceedings of the IEEE/CVF Conference on
  Computer Vision and Pattern Recognition}, 2021, pp. 4330--4339.

\bibitem{nguyen2017person}
D.~T. Nguyen, H.~G. Hong, K.~W. Kim, and K.~R. Park, ``Person recognition
  system based on a combination of body images from visible light and thermal
  cameras,'' \emph{Sensors}, vol.~17, no.~3, p. 605, 2017.

\bibitem{zhang2017alignedreid}
X.~Zhang, H.~Luo, X.~Fan, W.~Xiang, Y.~Sun, Q.~Xiao, W.~Jiang, C.~Zhang, and
  J.~Sun, ``Alignedreid: Surpassing human-level performance in person
  re-identification,'' \emph{arXiv preprint arXiv:1711.08184}, 2017.

\bibitem{sarfraz2018pose}
M.~S. Sarfraz, A.~Schumann, A.~Eberle, and R.~Stiefelhagen, ``A pose-sensitive
  embedding for person re-identification with expanded cross neighborhood
  re-ranking,'' in \emph{Proceedings of the IEEE conference on computer vision
  and pattern recognition}, 2018, pp. 420--429.

\bibitem{dai2018cross}
P.~Dai, R.~Ji, H.~Wang, Q.~Wu, and Y.~Huang, ``Cross-modality person
  re-identification with generative adversarial training.'' in \emph{IJCAI},
  vol.~1, no.~3, 2018, p.~6.

\bibitem{wang2019rgb}
G.~Wang, T.~Zhang, J.~Cheng, S.~Liu, Y.~Yang, and Z.~Hou, ``Rgb-infrared
  cross-modality person re-identification via joint pixel and feature
  alignment,'' in \emph{Proceedings of the IEEE/CVF International Conference on
  Computer Vision}, 2019, pp. 3623--3632.

\bibitem{zhong2020visible}
X.~Zhong, T.~Lu, W.~Huang, J.~Yuan, W.~Liu, and C.-W. Lin, ``Visible-infrared
  person re-identification via colorization-based siamese generative
  adversarial network,'' in \emph{Proceedings of the 2020 International
  Conference on Multimedia Retrieval}, 2020, pp. 421--427.

\bibitem{wang2020cross}
G.-A. Wang, T.~Zhang, Y.~Yang, J.~Cheng, J.~Chang, X.~Liang, and Z.-G. Hou,
  ``Cross-modality paired-images generation for rgb-infrared person
  re-identification,'' in \emph{Proceedings of the AAAI Conference on
  Artificial Intelligence}, vol.~34, no.~07, 2020, pp. 12\,144--12\,151.

\end{thebibliography}
	}

\end{document}